\documentclass{bmvc2k}
\usepackage{graphicx}
\usepackage{amsmath}
\usepackage{amssymb}
\usepackage{booktabs}
\usepackage{float}

\title{LLM-guided Instance-level Image Manipulation with Diffusion U-Net Cross-Attention Maps}

\addauthor{Andrey Palaev}{palaev@aidecisions.ai}{1}
\addauthor{Adil Khan}{a.m.khan@hull.ac.uk}{2}
\addauthor{Syed M. Ahsan Kazmi}{ahsan.kazmi@uwe.ac.uk}{3}

\addinstitution{
 AIDecisions, SW20 ODS, London, UK
}

\addinstitution{
 School of Computer Science, University of Hull, HU6 7RX, Hull, UK
}

\addinstitution{
Department of Computer Science,  University of the West of England, BS16 1QY, Bristol, UK
}

\runninghead{Palaev, Khan, Kazmi}{LLM-guided Instance-level Image Manipulation}

\def\etal{\emph{et al}\bmvaOneDot}

\begin{document}

\maketitle

\begin{abstract}
The advancement of text-to-image synthesis has introduced powerful generative models capable of creating realistic images from textual prompts. However, precise control over image attributes remains challenging, especially at the instance level. While existing methods offer some control through fine-tuning or auxiliary information, they often face limitations in flexibility and accuracy. To address these challenges, we propose a pipeline leveraging Large Language Models (LLMs), open-vocabulary detectors, cross-attention maps and intermediate activations of diffusion U-Net for instance-level image manipulation. Our method detects objects mentioned in the prompt and present in the generated image, enabling precise manipulation without extensive training or input masks. By incorporating cross-attention maps, our approach ensures coherence in manipulated images while controlling object positions. Our method enables precise manipulations at the instance level without fine-tuning or auxiliary information such as masks or bounding boxes. Code is available at \url{https://github.com/Palandr123/DiffusionU-NetLLM}
\end{abstract}

\section{Introduction}
\label{sec:intro}
Text-to-image synthesis, a field at the intersection of computer vision and natural language processing, tackles the challenge of generating visually realistic images from textual descriptions \cite{dhariwal2021diffusion, graikos2022diffusion, ho2020denoising, ramesh2021zero, Rombach_2022_CVPR}. This area holds immense potential for various applications, from revolutionizing human-computer interaction to creative content generation. The research community has recognized this significance, evidenced by the development of increasingly powerful text-to-image models such as Imagen \cite{saharia2022photorealistic}, DALL-E 3 \cite{Dalle3_2023} and Stable Diffusion 3 \cite{esser2024scaling}.\\
However, this field has some challenges. Current models often struggle to capture the full nuance of a text description, resulting in images that lack detail or contain nonsensical elements. Additionally, ensuring photorealism and semantic consistency across generated images remains a hurdle. Overcoming these obstacles is crucial, as it would pave the way for a future where humans can seamlessly communicate their creative vision through text, with machines acting as their capable artistic partners. Tackling these challenges can bridge the gap between human imagination and visual representation.\\
Among these challenges, a particularly important one is that creating the precise prompt to generate the desired image can be difficult. All desired image attributes should be conveyed through text, including those inherently complex or impossible to express accurately. Hence, designing a method for precise image editing is a crucial task in the field of text-to-image synthesis.\\
Previous research has tried to address the challenge of limited control in image editing. However, some methods rely on fine-tuning of pretrained models \cite{gal2023an, Kawar_2023_CVPR, Ruiz_2023_CVPR, Brooks_2023_CVPR} which is computationally expensive, require large amounts of data and may limit the range of edits. Other methods such as \cite{Tumanyan_2023_CVPR} inject diffusion features and self-attention maps to generate a new image while keeping details and appearance from the source one, limiting the range of possible edits. Some methods enable image editing in a zero-shot manner via editing cross-attention maps \cite{hertz2023prompttoprompt, epstein2023diffusion}, limiting only to object-type, not instance-level manipulations. Other methods take auxiliary information such as masks \cite{Avrahami_2022_CVPR, avrahami2023blendedlatent, mou2024dragondiffusion}, which is not always an option, or generate it  \cite{couairon2023diffedit} to better localize the region of interest, limiting the set of resulting edits.\\
Wu \etal \cite{wu2023self} proposed Self-correcting
LLM-controlled Diffusion (SLD) that automatically aligns the generated image with the user prompt. Firstly, it detects the objects described in the user prompt using a Large-Language Model (LLM) and open-vocabulary detector. Then, LLM finds inconsistencies between the user prompt and detection results and suggests the modification. Then, it performs latent operations to edit the image. This loop is repeated until LLM does not suggest any modifications. This method can be used not only for aligning the image with the prompt but also for image manipulation directly. However, the editing needs to be expressed through the text, limiting the manipulation precision.\\
To address the issues mentioned above, we propose a novel pipeline. Firstly, we utilize LLM and an open-vocabulary detector to detect the objects mentioned in the prompt and presented on the generated image in the same way as in \cite{wu2023self}. This enables instance-level manipulations without requiring any auxiliary information from the user. Then, we perform the instance-level manipulation specified by the user. In contrast to \cite{wu2023self} which performs latent operations using unsupervised segmentation, our method utilizes the guidance on cross-attention maps and intermediate activations of diffusion U-Net. This enables precise manipulation of such attributes as position while preserving the original image details. Hence, our pipeline enables to perform precise instance-level manipulations without fine-tuning or auxiliary information while ensuring the preservation of original appearances.
\section{Background \& Related work}
\label{sec:background}
This section provides the necessary background and overview of the related research. Section \ref{subsec:diffusion_models} describes the idea behind diffusion models, Section \ref{subsec:guidance} describes the guidance and Section \ref{subsec:image_editing} gives an overview of the related work on diffusion-based image editing.
\subsection{Diffusion models}
\label{subsec:diffusion_models}
Diffusion models use text prompts to generate high-res images from noise through sequential sampling \cite{ho2020denoising, kingma2021variational, song2021scorebased}. The aim is to reverse a time-dependent destructive process where noise corrupts data. A neural network $\epsilon_\theta$ estimates either the denoised image or the noise $\epsilon_t$ added to create the noisy image $z_t = \alpha_t x + \sigma_t * \epsilon_t$. Training involves minimizing the loss function:
\begin{equation}
L(\theta)=\mathbb{E}_{t\sim \mathcal{U}(1,T), \epsilon_t\sim\mathcal{N}(0, \textbf{I})}[||\epsilon_t-\epsilon_{\theta}(z_t; t, y)||_2^2],
\end{equation}
where $\epsilon_{\theta}$, often having a U-Net architecture with self- and cross-attention at different resolutions, incorporates conditioning signal $y$ such as text \cite{Rombach_2022_CVPR, saharia2022photorealistic}. Once the model is trained, the model can produce samples based on conditioning $y$ by setting the noise $z_T\sim\mathcal{N}(0, \textbf{I})$, then iteratively estimating the noise and updating the noisy image using techniques such as DDIM \cite{song2021denoising} or DDPM \cite{ho2020denoising}:
\begin{equation}
    \hat{\epsilon}_t=\epsilon_{\theta}(z_t; t, y), z_{t-1}=\text{update}(z_t, \hat{\epsilon}_t, t, t-1, \epsilon_{t-1})
\end{equation}
\subsection{Guidance}
\label{subsec:guidance}
Diffusion models offer post-training adjustment through guidance, involving the composition of score functions \cite{dhariwal2021diffusion, liu2022compositional, song2021scorebased}. Conditional samples can be generated using classifier guidance, combining unconditional score function $p(z_t)$ with classifier $p(y|z_t)$ as $p(z_t|y)\propto p(y|z_t)p(z_t)$ \cite{dhariwal2021diffusion, song2021scorebased}. Classifier guidance during sampling adjusts the estimated error term $\hat{\epsilon}_t$:
\begin{equation}
    \hat{\epsilon}_t = \epsilon_{\theta}(z_t; t, y)-s\sigma_t\nabla_{z_t} \log{p(y|z_t)},
\end{equation}
where $s$ sets guidance strength. This shifts sampling towards images the classifier considers more likely \cite{dhariwal2021diffusion}. Additionally, diffusion sampling can be guided using any energy function $g(z_t; t, y)$, not limited to classifier probabilities. Integrating such guidance yields high-quality text-to-image samples with low energy according to function $g$:
\begin{equation}
    \hat{\epsilon}_t = (1+s)\epsilon_{\theta}(z_t; t, y)-s\epsilon_{\theta}(z_t; t, \varnothing) + v\sigma_t\nabla_{z_t}g(z_t; t, y),
    \label{eq:guidance}
\end{equation}
where $v$ denotes an additional guidance weight for $g$.
\subsection{Diffusion-based Image editing}
\label{subsec:image_editing}
Image editing is a fundamental task in computer graphics, involving the manipulation of an input image by incorporating various additional elements, such as labels and reference images. Recent advances in text-to-image diffusion models expand their use in image editing tasks, including local and global edits.\\
Some methods attempted to solve this task by retraining or fine-tuning the diffusion model. For instance, InstructPix2Pix \cite{Brooks_2023_CVPR} generates an image editing dataset using GPT-3 \cite{brown2020language}, Stable Diffusion \cite{Rombach_2022_CVPR}, and Prompt-to-Prompt \cite{hertz2023prompttoprompt}, and then trains a diffusion model on this dataset to edit the image given the source image and editing prompt. Imagic \cite{Kawar_2023_CVPR} first optimizes the text embedding to the input image, then fine-tunes the diffusion model to further improve fidelity, and interpolates between the original and optimized embeddings to generate the resulting image. DreamBooth \cite{Ruiz_2023_CVPR} fine-tunes the diffusion model to reconstruct images of a specific object and objects of that type to generate new images of that object, given only 3-5 images of it. In comparison, Gal \etal \cite{gal2023an} proposed optimizing the vector embedding associated with the specific object, rather than the diffusion model, to minimize the reconstruction loss, given 3-5 images of that object. However, all these methods require retraining or fine-tuning of the diffusion model or optimization of the text embedding, which is computationally expensive and may limit the range of possible edits. In contrast, our method does not change the diffusion model weights and text embeddings by utilizing guidance.\\
Some methods attempted to perform image editing in a zero-shot manner. Tumanyan \etal \cite{Tumanyan_2023_CVPR} proposed injecting self-attention maps and features from diffusion U-Net during generation to preserve the original appearances and details. Prompt-to-Prompt \cite{hertz2023prompttoprompt} achieves certain types of image editing such as word addition, removal, and replacement by adding, removing, or replacing corresponding cross-attention maps during generation. Self-Guidance \cite{epstein2023diffusion} utilizes guidance with cross-attention maps and intermediate features of diffusion U-Net to manipulate attributes such as position, size, shape, and appearance. However, since these methods are based only on cross-attention maps, they can perform image editing only at the object-type level (i.e., manipulate all objects corresponding to the word, not a particular object), but not at the instance level. In contrast, our method can perform instance-level manipulations by extracting objects from the image using LLM and an open-vocabulary detector.\\
Blended Diffusion \cite{Avrahami_2022_CVPR} blends the CLIP-guided \cite{radford2021learning} latents with the original image at every diffusion image using the user-specified mask to achieve region-based image editing. Blended Latent Diffusion \cite{avrahami2023blendedlatent} further develops this idea by applying the same operation in the latent space rather than in the pixel space. DragonDiffusion \cite{mou2024dragondiffusion} manipulates the intermediate features of the diffusion model to perform different types of edits such as position change, resizing and object pasting, given the necessary editing masks. However, these approaches require mask specifying the region of interest as an input which is not always an option. DiffEdit \cite{couairon2023diffedit} automatically generates the editing mask based on the difference between the source and query prompts, then, at some diffusion steps, it blends the generation results from the query prompt with the source image. However, the generated mask is not precise. Such an approach also does not enable instance-level manipulations and limits the range of possible edits. In contrast, our method extracts the bounding boxes corresponding to every object mentioned in the prompt using LLM and an open-vocabulary detector. This enables to extract precise regions of interest without limiting the set of possible edits.\\
The integration of large language models (LLMs) with diffusion models has significantly advanced controlled image editing. LLM-grounded Diffusion (LMD) \cite{lian2023llmgrounded} enhances prompt comprehension through LLM-guided image layout generation. SUR-adapter \cite{zhong2023sur} aligns semantic representations of simple and complex prompts and transfers LLM knowledge for improved image generation. While these methods leverage LLMs for enhanced prompt interpretation, our approach utilizes LLMs and open-vocabulary detectors for precise instance-level editing.\\
Self-correcting LLM-controlled Diffusion (SLD) \cite{wu2023self} utilizes a different approach. Firstly, it extracts a set of objects from the prompt using LLM and detects them on the image. Then, LLM suggests necessary edits to make the image align with the prompt. Finally, it performs corresponding latent operations to edit the image. This loop is repeated until the image fully matches the prompt. We use the object extraction and detection part in our method since it enables to precisely locate the objects which should be manipulated or preserved without auxiliary information. However, the editing part is limited only to text-based image manipulation which is not precise. To enable more precise editing, we utilize the guidance based on the cross-attention maps on features from diffusion U-Net.\\
Custom Diffusion \cite{Kumari_2023_CVPR} introduces a fine-tuning framework that enables the customization of multiple concepts such as a specific dog and moongate within a single diffusion model. This method optimizes only key and value projection matrices in the cross-attention layers for the new text tokens against the user-provided concepts. In addition, the authors propose to add a subset of real images with similar captions to preserve the model’s ability to generate diverse images.\\
Similarly, Break-A-Scene \cite{avrahami2023break} explores the extraction of multiple concepts from a single image, enabling the decomposition of a complex scene into distinct elements that can be independently manipulated. This is achieved through a two-phase customization process that involves optimizing both textual embeddings and model weights, alongside employing a masked diffusion loss and a unique loss on cross-attention maps to prevent concept entanglement. This method excels in capturing the unique characteristics of each concept within the image, enabling diverse and flexible manipulations in subsequent generations. However, these two methods focus on general concept customization rather than instance-specific edits, i.e., they aim to generate completely new images with newly learned concepts rather than editing a given one. Our approach, by comparison, leverages precise instance-level manipulations, guided by the integration of LLMs and open-vocabulary detectors, enabling more targeted and flexible image editing without model retraining or fine-tuning.\\
ControlNet \cite{Zhang_2023_ICCV} provides a mechanism to control specific attributes in generated images by introducing conditional inputs without requiring fine-tuning of the entire model. This approach leverages pre-trained diffusion models and enables more fine-grained conditional generation. While this method provides a high level of control, it primarily focuses on guiding the generation process rather than directly editing an existing image. Our method, on the other hand, enables instance-level manipulations of objects within an image, without altering model weights, by integrating guidance techniques and leveraging LLMs and open-vocabulary detectors.\\
Dynamic Prompt Learning (DPL) \cite{wang2023dynamic} addresses the challenge of inaccurate cross-attention maps in diffusion models by optimizing word embeddings corresponding to the noun words. The authors propose to minimize three losses preventing the background leakage of the cross-attention, making them disjoint and focusing on the specific part of the image. This method enhances the precision of text-based edits by mitigating undesired alterations caused by cross-attention misalignments, enabling more accurate and controlled modifications. This method solves a related, but still a different problem, focusing on refining the interaction between text prompts and visual features to achieve high-quality edits.

\section{Methodology}
\label{sec:methodology}
\begin{figure}[ht]
\begin{center}
\includegraphics[width=1.0\textwidth]{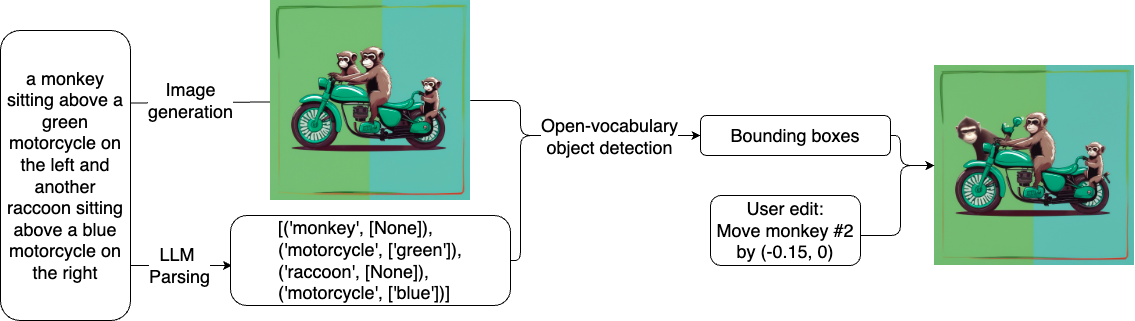}
\end{center}
\caption{Overview of our pipeline. Firstly, LLM parses the objects from the prompt. Then, an open-vocabulary object detector detects these objects on the image. Finally, the image is edited with the use of guidance. Note that this pipeline uses only pretrained models and does not require any training or fine-tuning.}
\label{fig:method}
\end{figure}
An overview of our method can be seen in Fig. \ref{fig:method}. Note that this pipeline uses only pretrained models and does not require any training or fine-tuning. Firstly, LLM parses the objects from the given prompt. Then, the open-vocabulary detector detects the parsed objects on the generated image. Then, given the user edit, we perform the image editing using guidance based on the cross-attention maps and features from the diffusion U-Net. Section \ref{subsec:detection} describes LLM parsing and open-vocabulary detection, and Section \ref{subsec: image_manipulation} describes the image editing with guidance.
\subsection{LLM-based object detection}
\label{subsec:detection}
In our method, LLM-based object detection extracts the objects mentioned in the prompt and present in the generated image. We do this in the same way as Wu \etal \cite{wu2023self}. Firstly, LLM extracts the objects mentioned in the prompt along with their attributes and quantities. Then, the open-vocabulary detector \cite{minderer2023scaling} detects the objects extracted during the previous step on the image. In contrast to methods such as Self-Guidance \cite{epstein2023diffusion}, which operate at the object level and cannot extract separate objects, these steps enable our method to precisely locate all objects of interest without requiring auxiliary information from the user, such as masks, unlike methods such as DragonDiffusion \cite{mou2024dragondiffusion}. Then, the image can be edited at the instance level by utilizing the obtained bounding boxes.
\subsection{Image editing with guidance}
\label{subsec: image_manipulation}
After obtaining the detection results, the user needs to specify which object needs to be manipulated. This enables more precise edits compared to methods that allow only text-based manipulations such as SLD \cite{wu2023self}. Then, image editing is performed using guidance based on cross-attention maps and features from diffusion U-Net. Only the position can be manipulated, but the method can be extended to other manipulations. Guidance has been shown to enable precise control over the image generation process \cite{bansal2024universal, ho2021classifierfree}, while recent research has demonstrated that cross-attention maps contain information about the object position and shape \cite{hertz2023prompttoprompt, epstein2023diffusion} and intermediate diffusion features contain information about object appearances \cite{Tumanyan_2023_CVPR, mou2024dragondiffusion}. Hence, this enables better control over the position while preserving appearances in the image, in contrast to methods such as SLD \cite{wu2023self} that directly inject objects into the latent vector, degrading image realism and fidelity.
\subsubsection{Position}
Given the original object bounding box $(x_1, y_1, x_2, y_2)$ and shift $(x, y)$, the position can be manipulated by using the following guidance term:
\begin{equation}
\begin{split}
g_{\text{position}}(o)=-\frac{1}{(x_2-x_1)(y_2-y_1)}\sum_{h,w}(\mathcal{A}_{h,w}*\mathcal{M}_{h,w}^{\text{target}})^2+\\
+\frac{1}{(x_2-x_1)(y_2-y_1)}\sum_{h,w}(\mathcal{A}_{h,w}*\mathcal{M}_{h,w}^{\text{orig}})^2,
\end{split}
\label{eq:g_position}
\end{equation}
where $\mathcal{A}_{k}$ is the cross-attention map corresponding to token $k$ obtained during the image editing, $\mathcal{M}^{\text{orig}}$ is the mask corresponding to the original bounding box, $\mathcal{M}^{\text{target}}$ is obtained by shifting $\mathcal{M}_{\text{orig}}$ by the shift $(x, y)$ and $*$ denotes element-wise multiplication. The first term aims to minimize the model’s focus on the original location, i.e., remove the object from there, while the second term aims to make the model focus on the target location, i.e., make the object appear at the target location.
\subsubsection{Object preservation}
For the rest of the objects that are not manipulated, we calculate the Mean Squared Error between the intermediate activations of diffusion U-Net obtained during the original generation denoted as $\Psi^{\text{orig}}$ and intermediate activations obtained during the manipulation denoted as $\Psi^{\text{target}}$:
\begin{equation}
g_{\text{preserve}}(o)=\frac{1}{(x_2-x_1)(y_2-y_1)}\sum_{h,w}(\Psi_{h,w}^{\text{orig}}*\mathcal{M}_{h,w}^{\text{orig}} - \Psi_{h,w}^{\text{target}}*\mathcal{M}_{h,w}^{\text{target}})^2
\label{eq:g_preserve}
\end{equation}
\subsubsection{Total guidance term}
Given the set of objects $\mathcal{O}$ and the manipulated object $o_k$, the total guidance term is as follows:
\begin{equation}
\begin{split}
    g = w_0\frac{1}{|O|-1}\sum_{o\neq o_k \in O} \frac{1}{|\Psi|}\sum_{i=0}^{|\Psi|}g_{\text{preserve}}(o)
    + w_1\frac{1}{|\mathcal{A}|}\sum_{i=0}^{|\mathcal{A}|}g_{\text{manipulation}}(o_k)
\end{split}
\label{eq:eq1}
\end{equation}
This guidance term is used to update the noise estimate according to Eq. \ref{eq:guidance}.
\section{Results \& Discussion}
\label{sec:results}
We selected the Gemma-7b instruction model \cite{gemmateam2024gemma} as our LLM, which has demonstrated superior performance compared to other state-of-the-art models like Mistral-7B-Instruct-v0.2 \cite{jiang2023mistral} and Llama 2 \cite{touvron2023llama}. For the open-vocabulary detector, we used OWLv2 \cite{minderer2023scaling}, the best performer in zero-shot open-vocabulary object detection. We tested our pipeline on the Stable Diffusion XL model \cite{podell2024sdxl}, a leading diffusion model. Our method applies the position guidance term from Eq. \ref{eq:g_position} to all cross-attention maps at the first upper block of the diffusion U-Net for precise position and shape information, and the preservation guidance term from Eq. \ref{eq:g_preserve} to the features of the third upper block for precise appearances and details.\\
We compared our method to Self-Guidance \cite{epstein2023diffusion}, which provides a method for manipulating position, although it provides manipulation only at the object-type level, not at the instance level. We also compared it to DragonDiffusion \cite{mou2024dragondiffusion}, which enables manipulating attributes such as position at the instance level but requires auxiliary information in the form of masks or bounding boxes. In contrast to Self-Guidance, our method enables instance-level manipulations. Unlike DragonDiffusion, our method does not require any auxiliary information. We have not compared our method to SLD \cite{wu2023self} since SLD enables specifying the manipulation using text only, which limits the specificity of the image editing. We have not compared our method to Custom Diffusion \cite{Kumari_2023_CVPR}, Break-A-Scene \cite{avrahami2023break}, and ControlNet \cite{Zhang_2023_ICCV} since the first two methods solve a different problem of multiple concept customization and ControlNet addresses the problem of controlled image generation rather than specific instance-level manipulation such as object repositioning. Other methods for image manipulation with diffusion models do not enable instance-level editing and do not enable manipulation of position.\\
Section \ref{subsec: position} highlights the precision of our approach in manipulating object positions at an instance level, demonstrating its superiority over current state-of-the-art methods such as Self-Guidance and DragonDiffusion. Section \ref{subsec: preservation} presents a comparative analysis of different preservation terms, showing the impact of using intermediate activations compared to cross-attention maps in maintaining appearance fidelity during manipulation. Designing a metric for evaluating image editing techniques is not yet a solved task, especially for methods that manipulate attributes such as position. Therefore, we used qualitative (i.e., visual) comparison for both experiments to directly visualize the results and assess our approach, similar to previous methods \cite{mou2024dragondiffusion, epstein2023diffusion, hertz2023prompttoprompt, avrahami2023blendedlatent, Ruiz_2023_CVPR}.

\begin{figure}[ht]
\begin{center}
\begin{tabular}{cccc}
Original image & Our method & Self-Guidance & DragonDiffusion \\
\includegraphics[width=.2\textwidth]{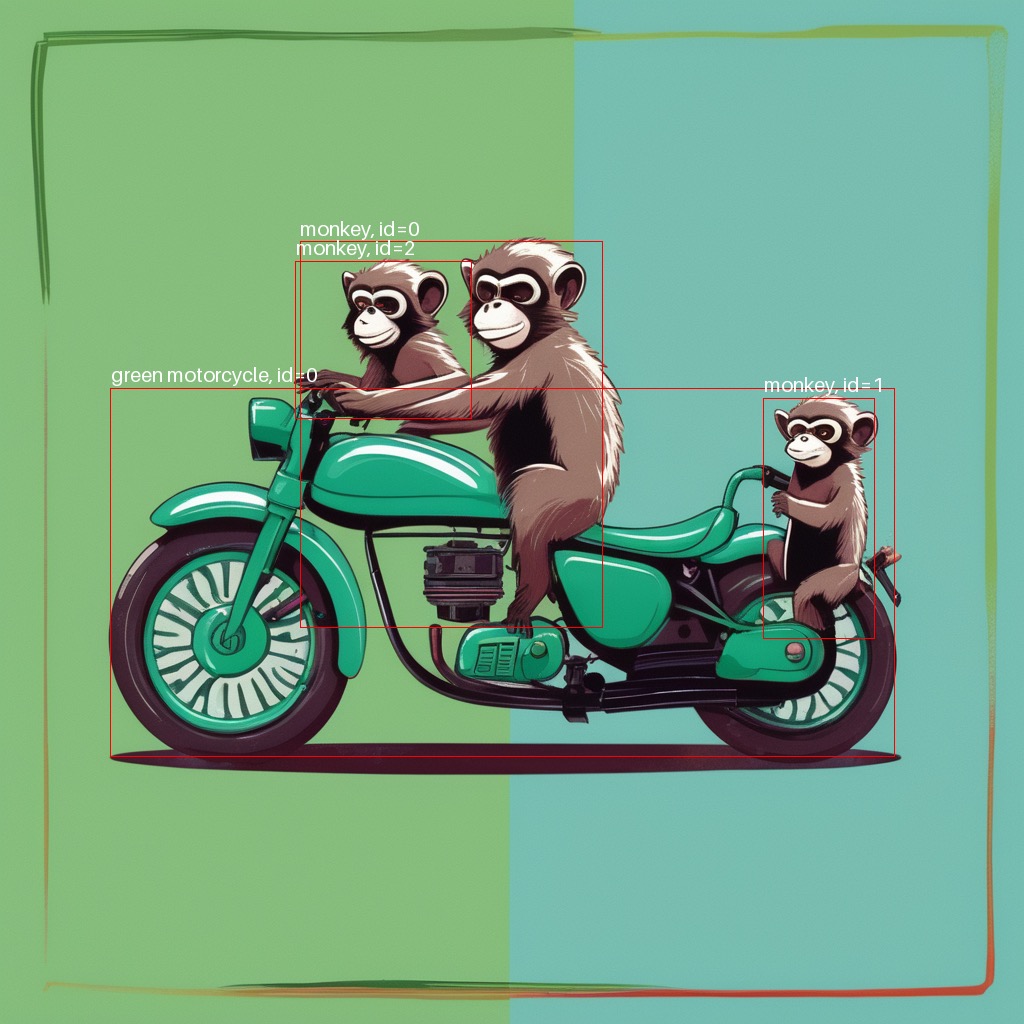}& \includegraphics[width=.2\textwidth]{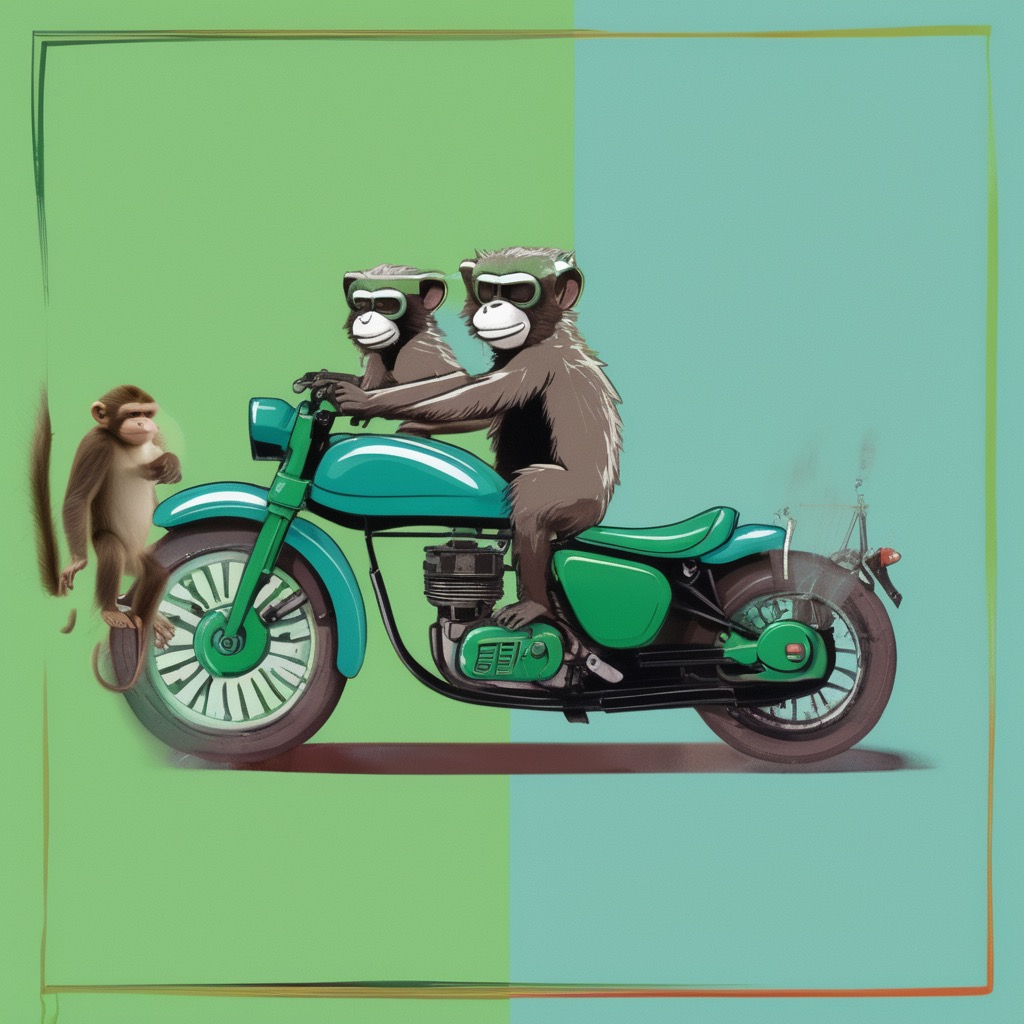} & \includegraphics[width=.2\textwidth]{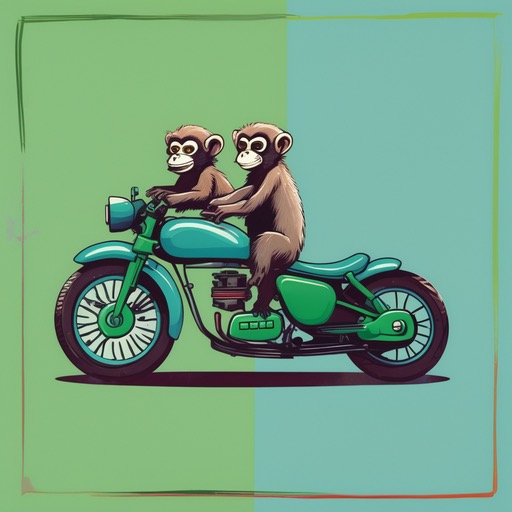} & \includegraphics[width=.2\textwidth]{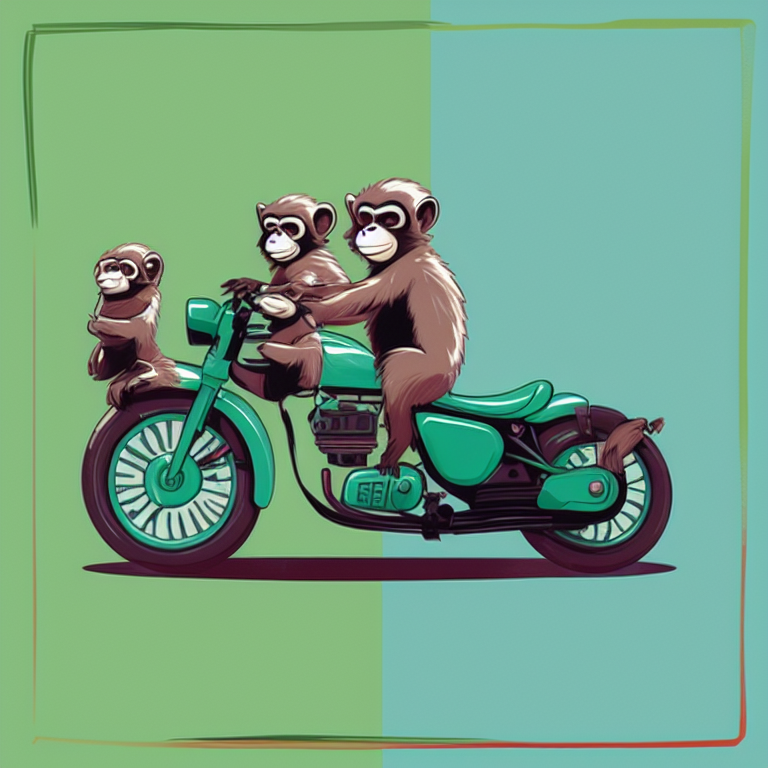}\\
\multicolumn{4}{c}{\shortstack{(a) Move monkey, id=1 (the rightmost) by $(-0.7, 0)$}}\\
\includegraphics[width=.2\textwidth]{images/monkeys_orig.jpeg}& \includegraphics[width=.2\textwidth]{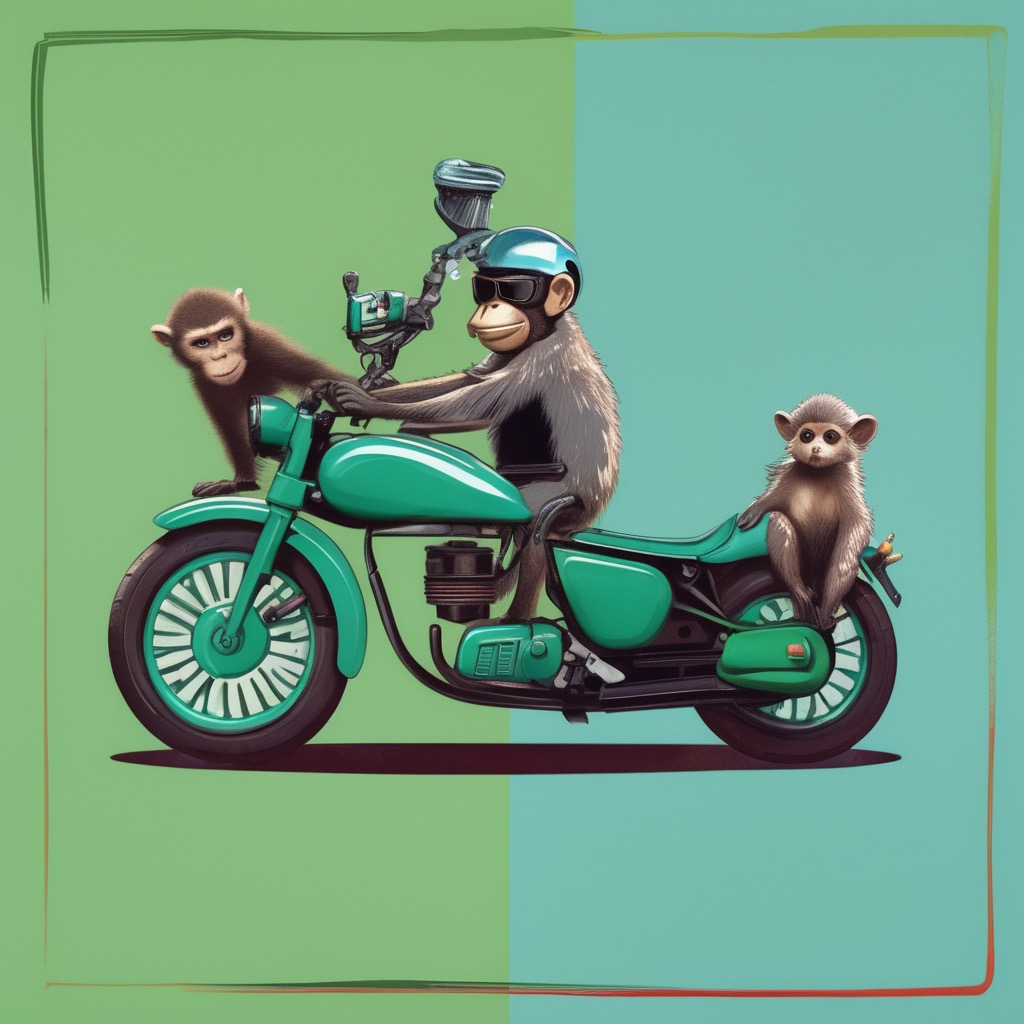} & \includegraphics[width=.2\textwidth]{images/sg_position1.jpeg} & \includegraphics[width=.2\textwidth]{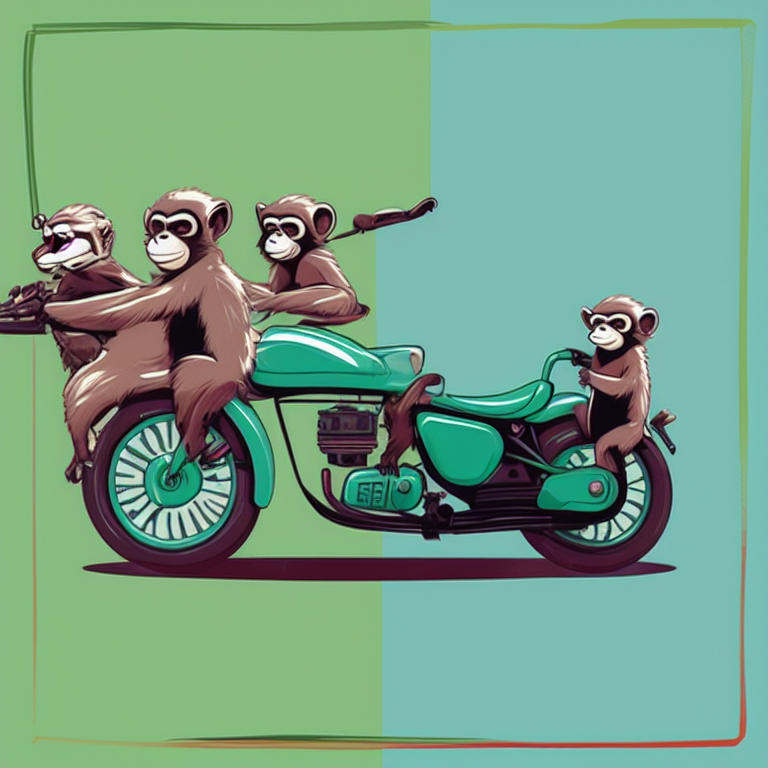}\\
\multicolumn{4}{c}{\shortstack{(b) Move monkey, id=0 (in the center) by $(-0.25, 0)$}}\\
\includegraphics[width=.2\textwidth]{images/monkeys_orig.jpeg}& \includegraphics[width=.2\textwidth]{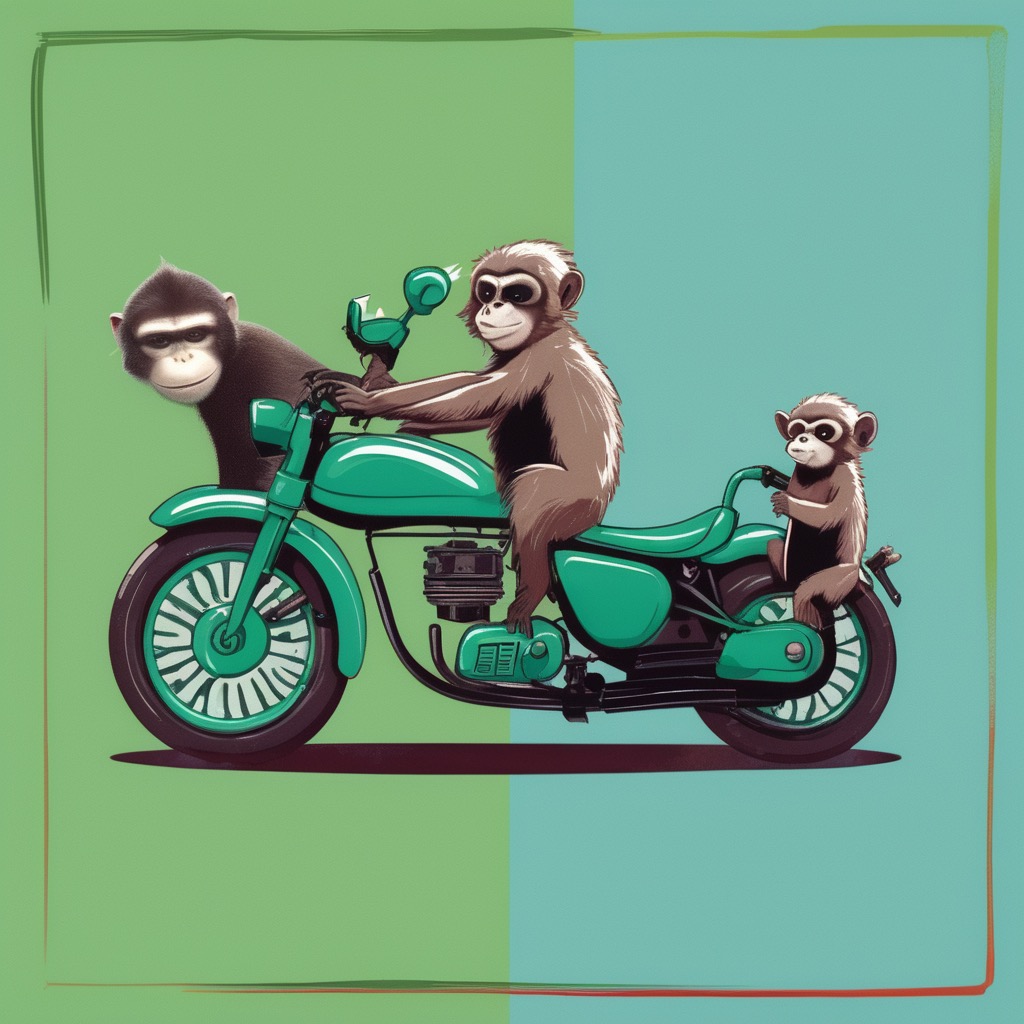} & \includegraphics[width=.2\textwidth]{images/sg_position1.jpeg} & \includegraphics[width=.2\textwidth]{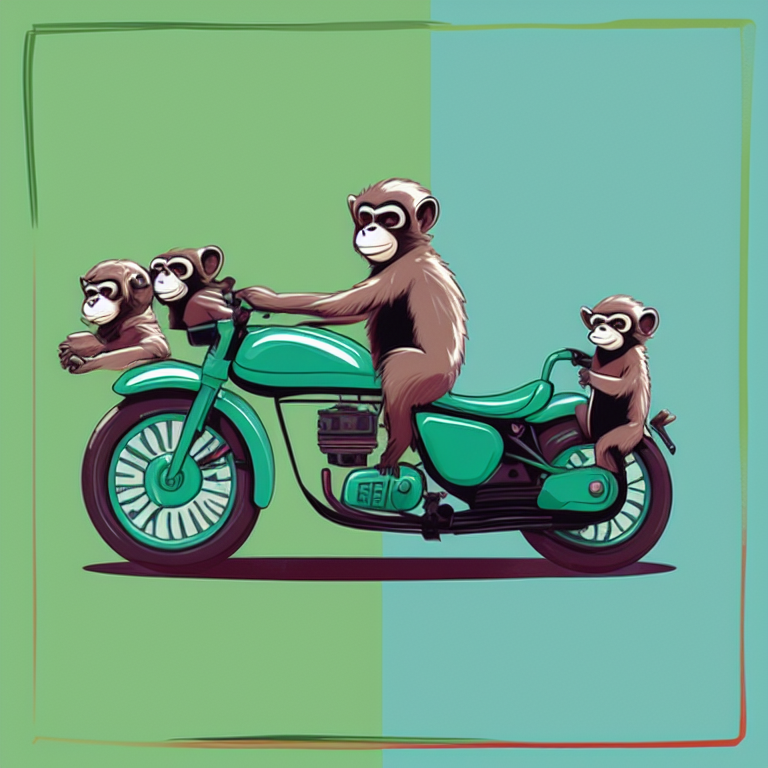}\\
\multicolumn{4}{c}{\shortstack{(c) Move monkey, id=2 (the leftmost) by $(-0.25, 0)$}}\\
\includegraphics[width=.2\textwidth]{images/monkeys_orig.jpeg}& \includegraphics[width=.2\textwidth]{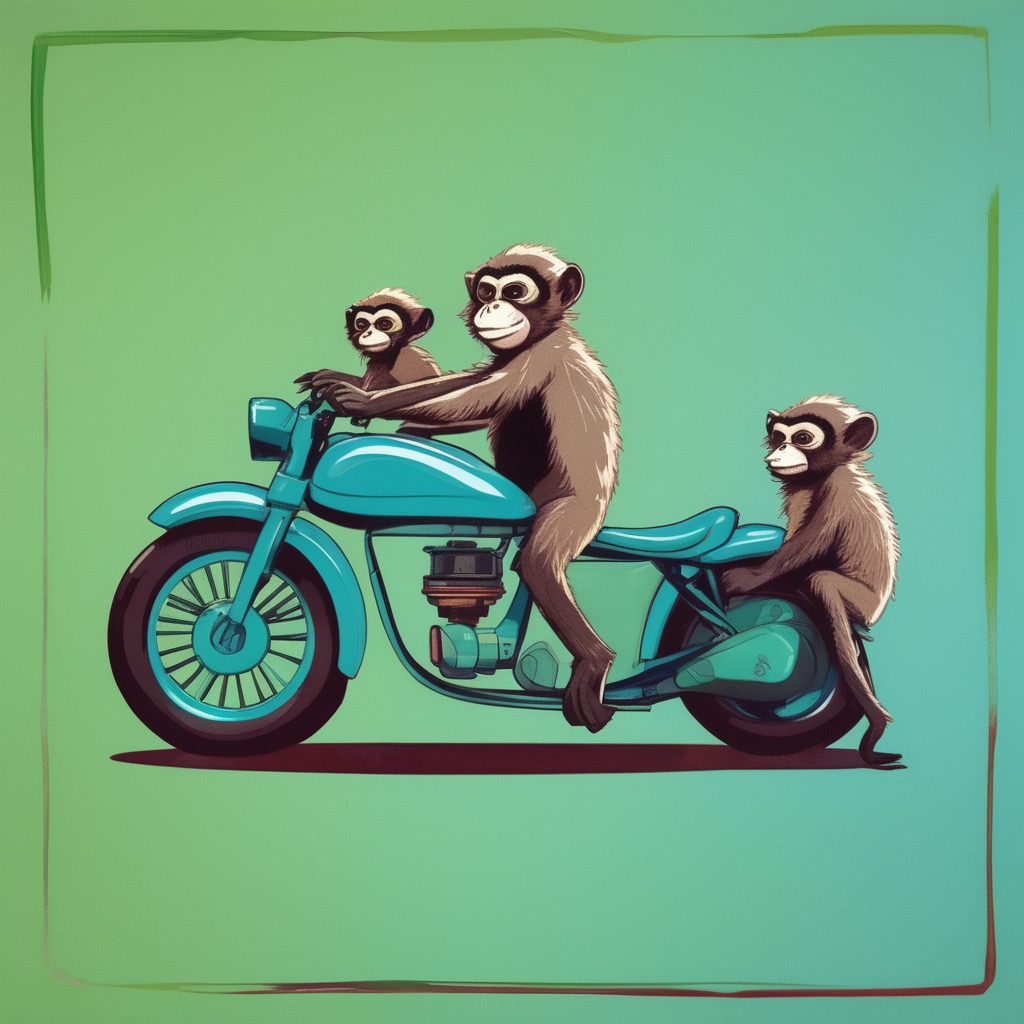} & \includegraphics[width=.2\textwidth]{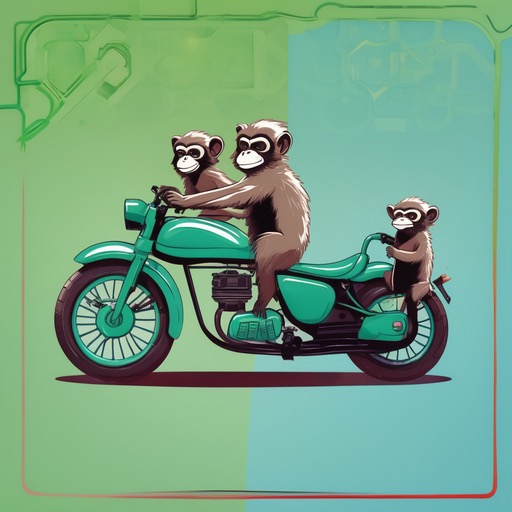} & \includegraphics[width=.2\textwidth]{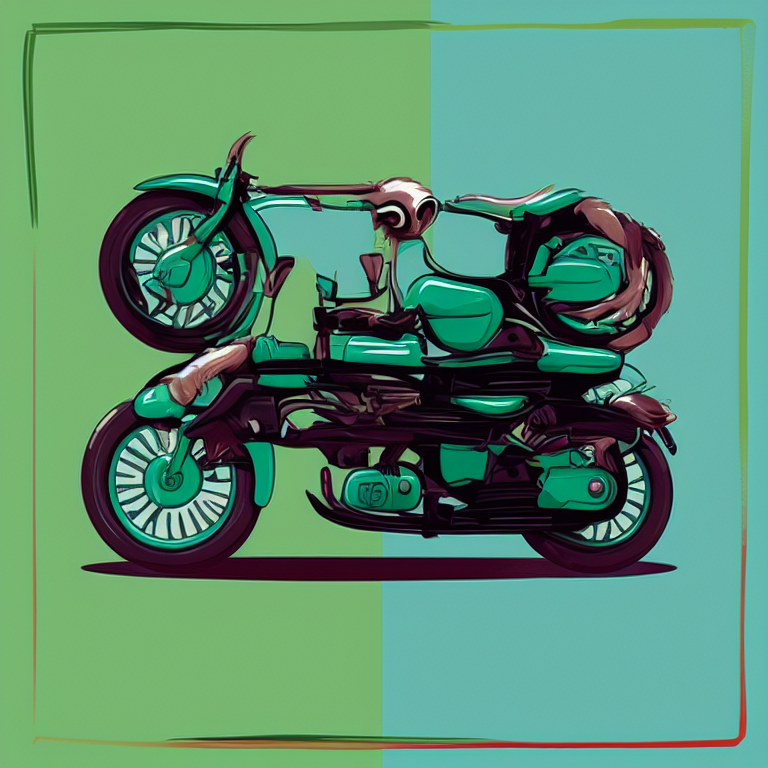}\\
\multicolumn{4}{c}{\shortstack{(d) Move green motorcycle by $(0, -0.25)$}}
\end{tabular}
\end{center}
\caption{Examples of the position manipulations. Coordinates shift is represented by (x, y)}
\label{fig:position}
\end{figure}

\subsection{Position}
\label{subsec: position}
Examples of position manipulation are shown in Fig. \ref{fig:position}. Our method achieves precise position manipulation while largely preserving the appearance fidelity of manipulated objects. Notably, it can manipulate specific instances of objects, unlike Self-Guidance \cite{epstein2023diffusion}, which controls object types rather than individual instances. For example, our method can manipulate individual monkeys while preserving their distinct appearances, whereas Self-Guidance gives the same result for any monkey, unable to differentiate between instances of the same object type. Compared to DragonDiffusion \cite{mou2024dragondiffusion}, our method maintains fidelity and realism in manipulated images, especially where objects are repositioned. While DragonDiffusion may preserve appearances more precisely, our method excels in maintaining realism and fidelity, crucial for applications requiring realistic modifications. Additional examples and quantitative evaluation can be found in the supplementary materials.\\
Despite these strengths, our method has areas for improvement. While it generally preserves appearances well, there are occasional deviations, such as color shifts in objects like monkeys or minor changes in motorcycle details. Additionally, the appearance of the moved object changes completely because the position term in Eq. \ref{eq:g_position} uses cross-attention maps that contain only general location and shape information, not appearances.\\
Another issue is the need for careful hyperparameter selection for each manipulation, which can be challenging in practical applications. The weights require tuning, as a combination that works well for one object may not yield satisfactory results for another.\\
Moreover, our method, like Self-Guidance and DragonDiffusion, struggles with manipulating large objects. For example, attempting to reposition the motorcycle in Figure \ref{fig:position} resulted in no displacement, highlighting a limitation in handling substantial changes without compromising image fidelity. Although DragonDiffusion managed to move the motorcycle upward, it failed to remove it from its original location.

\begin{figure}[!ht]
\begin{center}
\begin{tabular}{ccc}
Original image & Cross-attention maps & Intermediate features\\
\includegraphics[width=.2\textwidth]{images/monkeys_orig.jpeg}& \includegraphics[width=.2\textwidth]{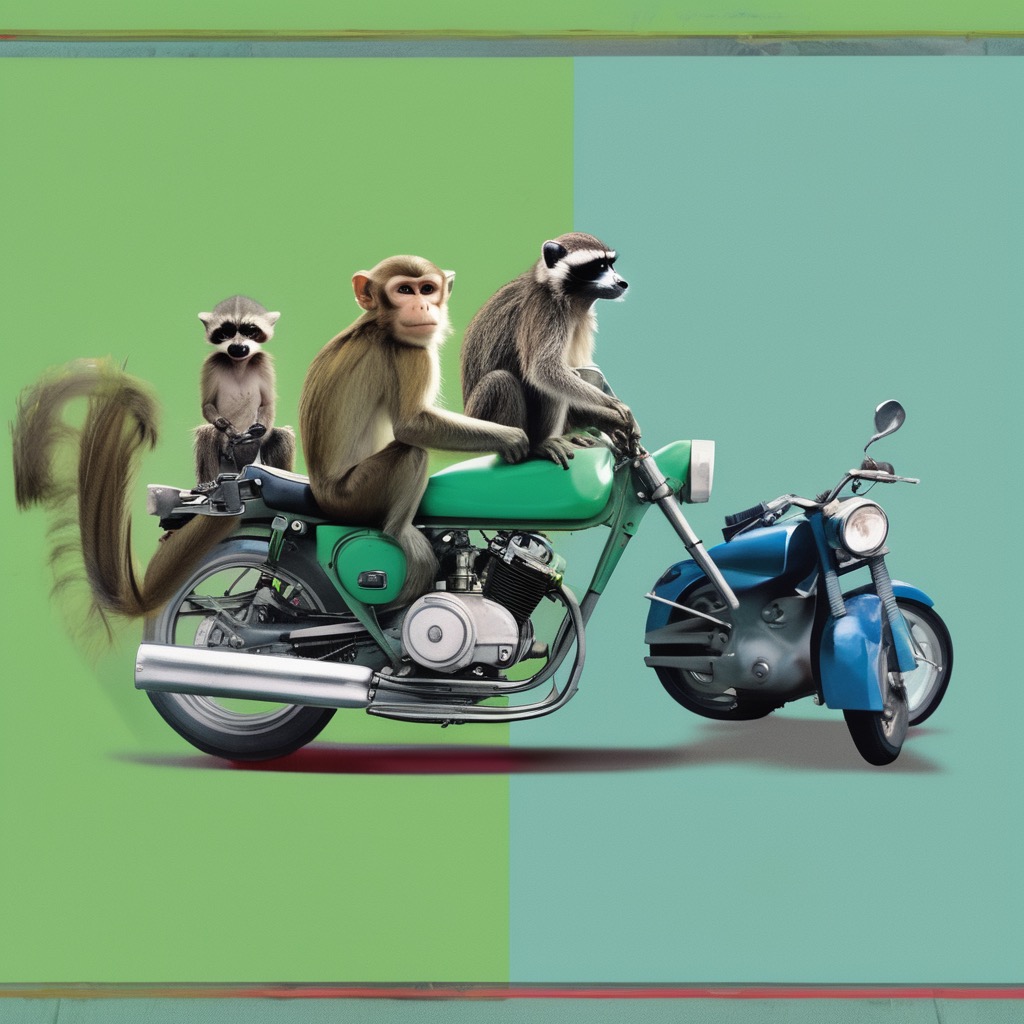} & \includegraphics[width=.2\textwidth]{images/monkeys1_1.jpeg} \\
\multicolumn{3}{c}{\shortstack{(a) Move monkey, id=1 (the rightmost) by $(-0.7, 0)$}}\\
\includegraphics[width=.2\textwidth]{images/monkeys_orig.jpeg}& \includegraphics[width=.2\textwidth]{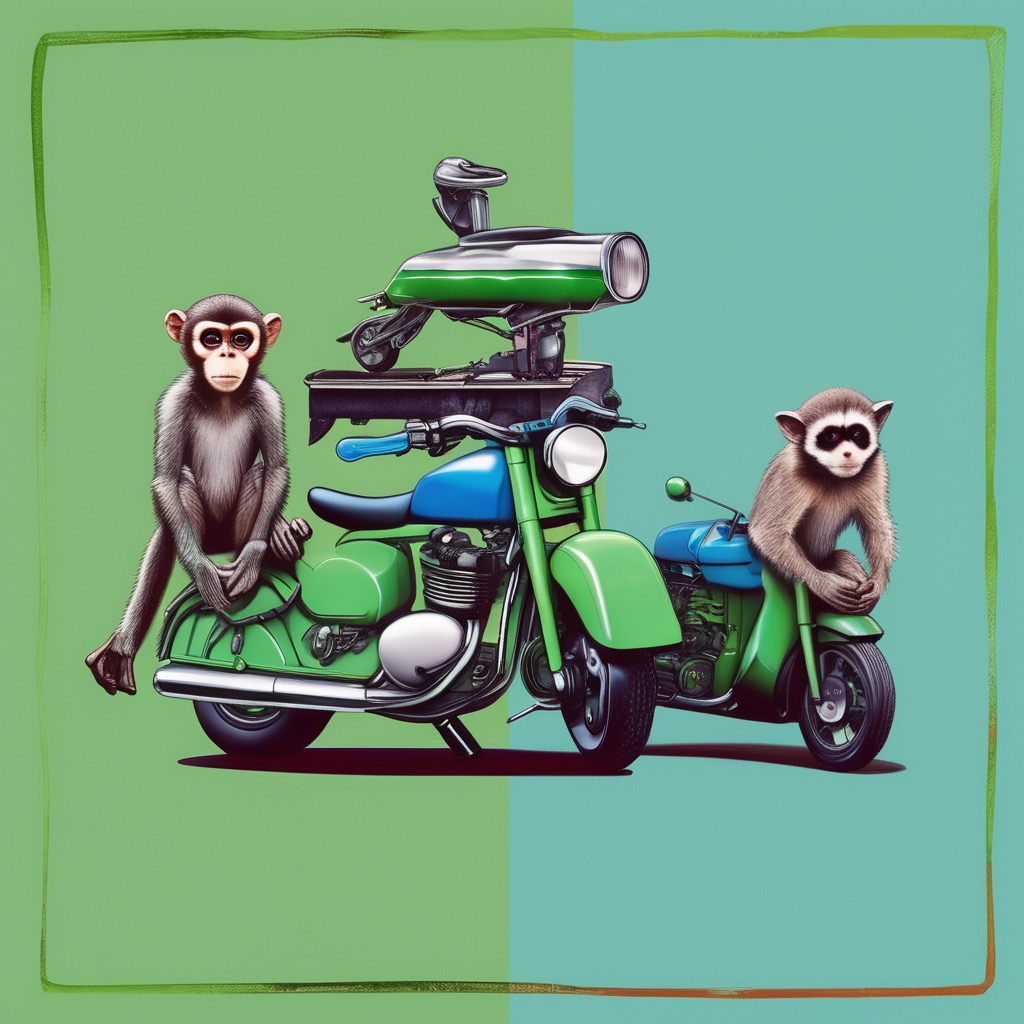} & \includegraphics[width=.2\textwidth]{images/monkeys2_1.jpeg}\\
\multicolumn{3}{c}{\shortstack{(b) Move monkey, id=0 (in the center) by $(-0.25, 0)$}}\\
\includegraphics[width=.2\textwidth]{images/monkeys_orig.jpeg}& \includegraphics[width=.2\textwidth]{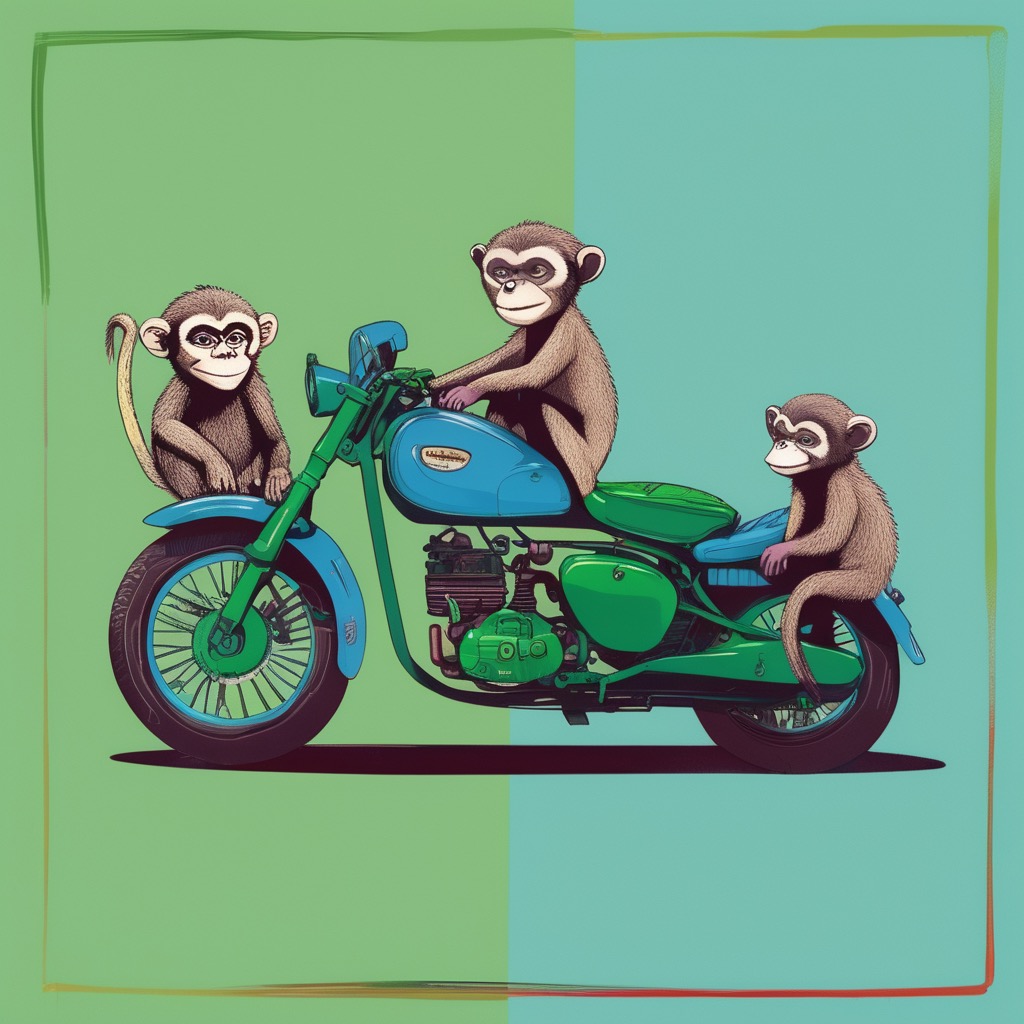} & \includegraphics[width=.2\textwidth]{images/monkeys3_1.jpeg} \\
\multicolumn{3}{c}{\shortstack{(c) Move monkey, id=2 (the leftmost) by $(-0.25, 0)$}}\\
\includegraphics[width=.2\textwidth]{images/monkeys_orig.jpeg}& \includegraphics[width=.2\textwidth]{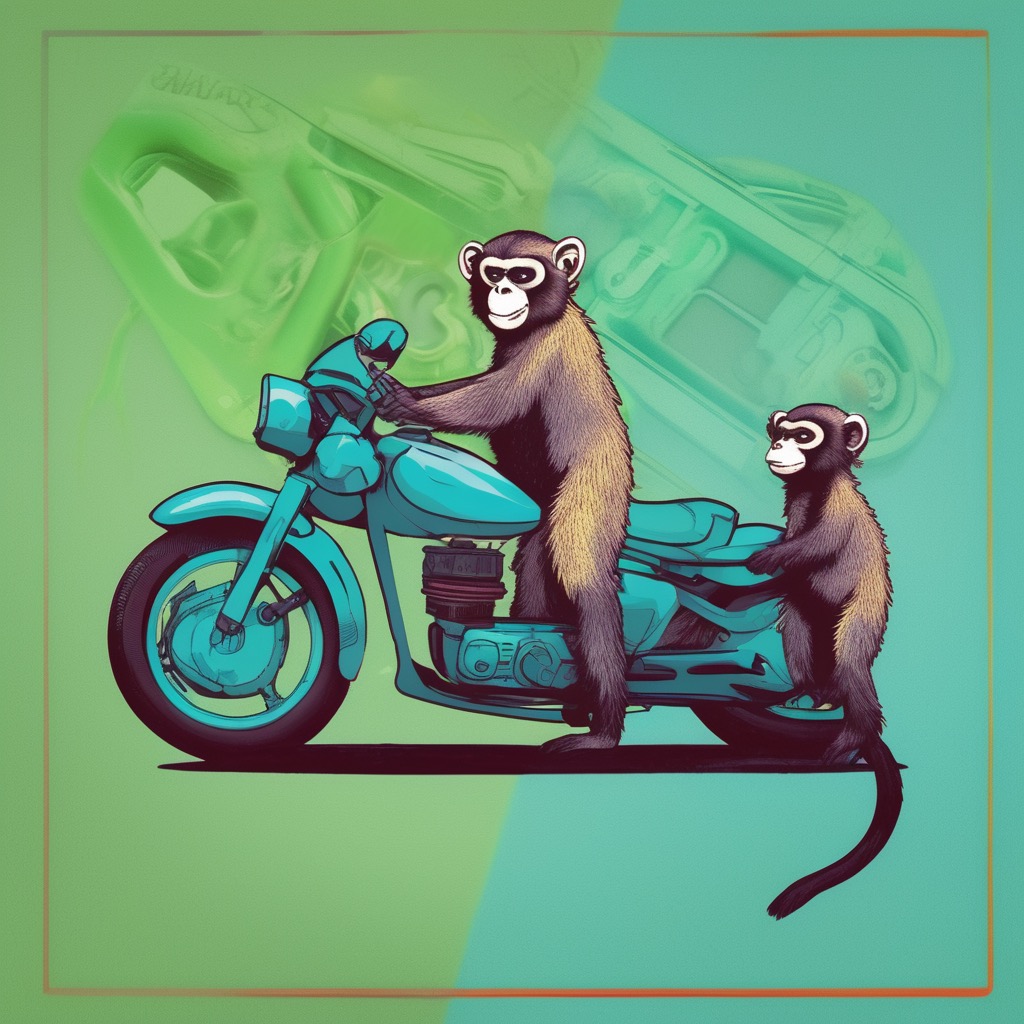} & \includegraphics[width=.2\textwidth]{images/motorcycle1_1.jpeg} \\
\multicolumn{3}{c}{\shortstack{(d) Move green motorcycle by $(0, -0.25)$}}

\end{tabular}
\end{center}
\caption{Comparison of different preservation terms in our method. Coordinates shift is represented by (x, y).}
\label{fig:preserve_comparison}
\end{figure}

\subsection{Ablation study on different preservation terms}
\label{subsec: preservation}

We compared two preservation terms in Eq. \ref{eq:g_preserve}: one using cross-attention maps as the position manipulation term in Eq. \ref{eq:g_position} and one using intermediate activations of diffusion U-Net. As shown in Fig. \ref{fig:preserve_comparison}, cross-attention maps preserve only general location and shape, significantly altering appearances. In contrast, intermediate activations of diffusion U-Net retain both position, shape, and appearances.

\section{Conclusion \& Future Work}
\label{sec:conclusion}
In this paper, we presented a pipeline for instance-level image manipulation. Our method detects objects mentioned in the prompt and present in the generated image by leveraging LLMs and open-vocabulary detectors, facilitating precise instance level control without the need for expensive fine-tuning or auxiliary information such as input masks. Additionally, our method maintains the image’s appearance, ensuring coherence.\\
Future work will focus on making our approach less sensitive to the choice of hyperparameters. Eliminating the need to tune hyperparameters for each manipulation will greatly enhance user convenience. In addition, we will focus on improving the preservation of the manipulated object's appearance during manipulation and on improving the position manipulation of large objects. Our method and previous methods do not solve these problems, and addressing them will further enhance the precision of image editing approaches. Finally, we will explore combining our method with DPL \cite{wang2023dynamic}, as it improves the precision of cross-attention maps, potentially enhancing the editing component of our pipeline.
\bibliography{bmvc_final}
\end{document}